%% file: main.tex
\documentclass[runningheads]{llncs}

\input{packages}

\input{macro}

\begin{document}

\title{VERSA: Verified Event Data Format for Reliable Soccer Analytics}

\titlerunning{Verified Event Data Format}

\institute{}

\author{
    Geonhee Jo\inst{1}\orcidlink{0009-0001-3717-7040} \and
    Mingu Kang\inst{1}\orcidlink{0009-0002-5724-3217} \and
    Kangmin Lee\inst{1}\orcidlink{0009-0007-2861-7386} \and
    Minho Lee\inst{2}\orcidlink{0000-0001-6762-5900} \and
    Pascal Bauer\inst{2}\orcidlink{0000-0001-8613-6635} \and
    Sang-Ki Ko\inst{1}\orcidlink{0000-0002-5406-5104}\corr
}

\authorrunning{Jo et al.}

\institute{
    University of Seoul, Seoul, Republic of Korea \\
    \email{\{geonhee,alsrn7182,rkdalsfpdl03,sangkiko\}@uos.ac.kr}
\and
Saarland University, Saarbrücken, Germany \\
    \email{\{minho.lee,pascal.bauer\}@uni-saarland.de}
}

\maketitle              

\begin{abstract}

Event stream data is a critical resource for fine-grained analysis across various domains, including financial transactions, system operations, and sports. In sports, it is actively used for fine-grained analyses such as quantifying player contributions and identifying tactical patterns. 
However, the reliability of these models is fundamentally limited by inherent data quality issues that cause logical inconsistencies (e.g., incorrect event ordering or missing events). 
To this end, this study proposes \versa (Verified Event Data Format for Reliable Soccer Analytics), a systematic verification framework that ensures the integrity of event stream data within the soccer domain. \versa is based on a state-transition model that defines valid event sequences, thereby enabling the automatic detection and correction of anomalous patterns within the event stream data. 
Notably, our examination of event data from the K League 1 (2024 season), provided by \bepro, detected that 18.81\% of all recorded events exhibited logical inconsistencies. 
Addressing such integrity issues, our experiments demonstrate that \versa significantly enhances cross-provider consistency, ensuring stable and unified data representation across heterogeneous sources. 
Furthermore, we demonstrate that data refined by \versa significantly improves the robustness and performance of a downstream task called \vaep~\cite{Decroos19vaep}, which evaluates player contributions
These results highlight that the verification process is highly effective in increasing the reliability of data-driven analysis. 

\keywords{Automated Data Verification \and Soccer Event Data \and Performance Analytics in Sports}

\end{abstract}

\section{Introduction}

Event-stream data consists of irregularly timed, context-dependent sequences that complicate sequential pattern analysis in finance, system operations, and healthcare. In team sports, where the importance of data analysis has grown, event streams have become a central component of modern data-driven analysis. In soccer, an event stream contains detailed records of on-ball actions (e.g., passes and shots), specifying the player, location, timestamp, and action type. This structured representation enables the quantitative evaluation of player performance through traditional metrics such as Expected Goals (xG)~\cite{Green2012}. Furthermore, it facilitates the development of predictive models that estimate outcomes, such as scoring a goal~\cite{Decroos19vaep,Liu20GIM} or recovering possession~\cite{Robberechts19VPEP,korean2025express,korean2025expressv2}. 

However, the analytical reliability of these models is fundamentally constrained by the quality of the underlying data. This degradation in data quality stems from several factors. 
First, event stream data is typically collected through manual annotation of match videos, where human operators record events based on subjective interpretation, inherent ambiguity, and human error.
Second, heterogeneous schema and semantics among major providers (e.g., \bepro and \sbomb) critically hinder the data unification necessary for cross-league analyses, since each provider typically covers different leagues (e.g., \bepro for the K League and \sbomb for the J League).

The ramifications of these quality issues severely undermine the reliability of analysis~\cite{anderson2013numbers}. For instance, when a Defensive block event is recorded before the Shot it was intended to stop, the event sequence violates temporal causality where an effect precedes its cause. This systematically misleads any model attempting to learn causal relationships or perform credit assignment from the event stream data.
This temporal inconsistency is analogous to critical errors in other domains; for instance, an approval for a fraudulent transaction being recorded before the transaction attempt in financial data would similarly render the analysis meaningless.
Our examination of event data from the K League 1 (2024 season), provided by \bepro, detected that 18.81\% of all recorded events exhibit logical inconsistencies.

These findings highlight that the reliability of analytical insights depends more on the robustness of the underlying data than on the model’s predictive accuracy. Analytical outcomes derived from robust data should produce consistent results across different parts of the season, time periods, or data providers, and should not change because of small timestamp errors or harmless preprocessing differences \cite{davis2024methodology,Jesse25Robustness,linke2018validation}. 
In practice, player valuation models such as \vaep~\cite{Decroos19vaep} are highly sensitive to these issues—missing or misordered events can distort action values and lead to unstable player ratings. Ensuring robustness is therefore essential for generating trustworthy and reproducible insights in soccer analytics.

To address these challenges, we propose \versa (Verified Event Data Format for Reliable Soccer Analytics), a systematic verification framework that ensures the logical and structural integrity of event stream data in soccer. \versa formalizes domain-specific knowledge as a state-transition model, which defines valid event sequences and enforces consistency rules. This formalization enables automatic detection and correction of anomalous event patterns, thereby enhancing the overall reliability of event-based analytics.

The key contributions of this work are as follows:
\begin{itemize}
    \item We propose a formal verification framework based on a state-transition model~\cite{bestelmeyer2017state} that automatically detects and corrects logical inconsistencies in soccer event streams.
    \item We introduce a unified event representation using \versa to integrate heterogeneous datasets across data providers and quantitatively validate enhanced data consistency and robustness on real-world datasets.
    \item We demonstrate that refining data with \versa improves the predictive performance and reliability of downstream tasks.
\end{itemize}

\section{Related Work}

\subsection{Sports Analytics in Soccer}

Soccer matches are typically recorded in two types of data: event data and tracking data~\cite{pappalardo2019public,bassek2025integrated}. Event data includes discrete actions such as passes and shots, which are primarily collected through manual annotation. Tracking data consist of the coordinates of all players and the ball, which are collected through GPS-based tracking and video-based systems. The advent of soccer event stream data has long been a central focus of soccer analytics, providing deeper insights into player contributions beyond the traditional metrics such as goals and assists that have been used for decades~\cite{decroos2020soccer}. Foundational player evaluation models include Expected Goals (xG)~\cite{Green2012}, which predicts the likelihood of a shot resulting in a goal based on contextual factors such as its location and angle. In addition, the Expected Threat (xT)~\cite{Singh19xT} model, which quantifies the value of moving the ball to specific positions on the field, is also widely used. 
However, these early models primarily focus on individual actions at a specific moment, limiting their ability to capture dynamic context such as the preceding sequence of events or opponent pressure.
To address this, recent studies have enhanced evaluations by integrating richer context and sequential dependencies into their models~\cite{Decroos19vaep,Liu20GIM,Robberechts19VPEP,korean2025express,korean2025expressv2}. 

\subsection{Data Quality and Reliability in Soccer Analytics}

Prior research has primarily addressed the quality of event stream data through three key dimensions: \textit{Synchronization}, \textit{Standardization}, and \textit{Completeness}.

First, \textit{Synchronization} refers to how well the spatiotemporal attributes of recorded events align with the ground truth of a match. As event streams are often collected through manual annotation, temporal discrepancies can occur between actual and recorded timestamps, which can distort any analysis that relies on spatiotemporal context. To address this, synchronization studies have been proposed to correct temporal misalignments between event data and tracking data~\cite{Anzer22xPass,Davis2024ETSY,hyunsung2025ELASTIC}.

Second, \textit{Standardization} refers to the uniformity of event definitions and data formats across providers. For example, different providers may use distinct terms for similar actions (e.g., Error in \bepro vs. Miscontrol in \sbomb), which complicates cross-league analyses. To address this, recent studies have proposed unifying data formats across various sources by mapping them to a common format~\cite{Decroos19vaep,atomic2020peter,gabriel2025CDF}.

Third, \textit{Completeness} refers to the absence of missing information in the dataset. There often exist implicit actions that are not explicitly recorded in matches. For instance, if a teammate gains possession immediately after an opponent's pass, an unrecorded `Interception' must have taken place. Such omissions can undermine the fairness of player contribution assessments. To address this, event-reconstruction frameworks introduce mechanisms that infer missing implicit events (e.g., based on possession-change)~\cite{Decroos19vaep,atomic2020peter}.

However, existing studies have not systematically validated the intrinsic sequential grammar of the event stream, leaving a critical gap in data quality assurance. Addressing these inconsistencies is indispensable because analytical reliability depends not only on data quality but also on robustness against such perturbations, preventing unstable or misleading insights~\cite{davis2024methodology,Jesse25Robustness}.

\section{Automatic Verification of Soccer Event Data}

\begin{figure*}[ht!]
    \centering
    \includegraphics[width=\textwidth]{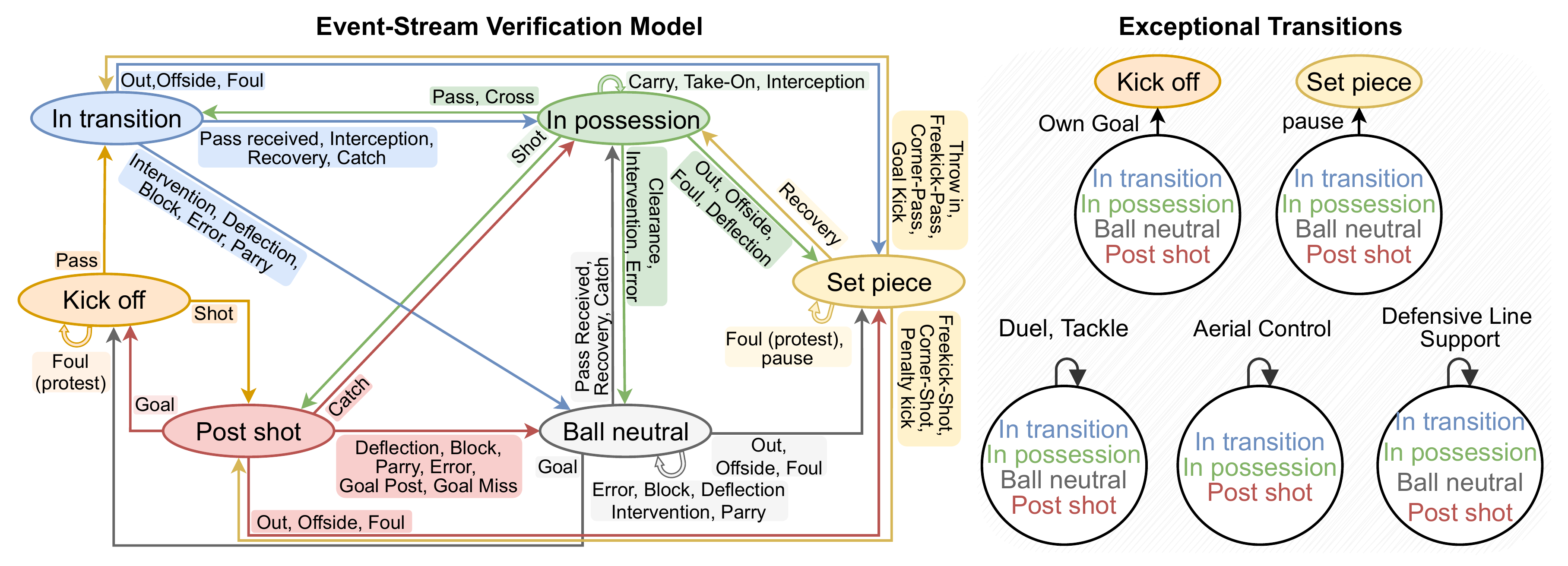}
    \caption{Event stream verification model using state-transition diagram}
    \label{fig:verification}
\end{figure*}

\label{sec: automatic verification}
Our methodology introduces \versa (Verified Event Data Format for Reliable Soccer Analytics), an {\em event-stream verification framework} designed to address data integrity issues systematically. 
The core of \versa is a formal state-transition model designed to efficiently validate and correct the complex event sequences in soccer. 
We first introduce the formal state-transition model and then explain correction mechanisms that detect and resolve logical inconsistencies.

\subsection{State Transition Model}
\label{sec:state_transition_model}
The core of \versa is a formal state-transition model designed to capture the logical flow of a soccer match. We implemented this model using the Transition library\footnote{\url{https://github.com/pytransitions/transitions}}. 

To construct this model, we first conceptualize the game as a sequence of discrete states. A state represents the logical context of the game at a specific moment (e.g., whether a team is in possession or the ball is in motion). We define a total of six mutually exclusive states to formalize the match flow: 

\begin{itemize}[leftmargin=*]
\item{\bf Kick-off}: Before the first pass is played from the center circle at the start of the match or after a goal.
\item{\bf In transition}: The ball is not possessed by a specific team or player and is in motion between players with a clear intent.
\item{\bf In possession}: A player has full control of the ball and can execute intentional actions for attacking, such as passing, dribbling, or shooting.
\item{\bf Ball neutral}: The ball is out of control of any player or team, typically occurring due to rebounds, blocks, or deflections.
\item{\bf Set piece}: The ball is reintroduced into play from a dead-ball situation such as a free-kick, corner-kick, or throw-in.
\item{\bf Post shot}: After a shot has been taken, transitioning based on the shot outcome (Goal, Catch, Block, or Out).
\end{itemize}

Second, we define the transitions between these states. Each action type is defined as a Transition Label. When an event occurs, its label triggers a transition from the source state to a predefined target state. Here, every Transition Label must correspond to only one specific state transition. For example, a `Pass' event always transitions the current state to `In transition'. 
Furthermore, the model defines exceptional transitions for events that do not follow standard transitions. 
Events that do not directly affect state transitions are processed using a self-loop mechanism, meaning that the current state is maintained (see the Exceptional Transitions illustrated at the right of Figure~\ref{fig:verification}). 
For example, off-ball events such as `Duel' are processed using a self-loop mechanism, maintaining the current state.
`Own Goal' is an event that can occur from any state (excluding `Set piece') and immediately transitions the current state to `Kick-off'.
The complete set of valid states and transitions is illustrated in the state-transition diagram in Figure~\ref{fig:verification}. 

Third, we define explicit transition conditions. While action types provide the primary triggers for state transitions, relying solely on them is insufficient to capture all the logical constraints inherent in the game. Specifically, certain transitions depend not only on the action type but also on associated attributes (e.g., outcome, location) or contextual information. For example, a `Pass Received' event following a failed `Pass' violates the logical flow: the transition from `In transition' to `In possession' should only occur if the preceding `Pass' was successful. Similarly, a `Carry' event may be required when a player performs two consecutive actions with spatial separation exceeding a threshold  (e.g., 3 meters). To address these issues, we define explicit transition conditions that combine action types with relevant event attributes and contextual information.
Based on this structure, \versa verifies events sequentially as described in Algorithm~\ref{alg:pseudo_code}. 
A transition is allowed only when a valid target state exists and the associated conditions are satisfied. Conversely, if an event attempts a transition that violates these rules, it is flagged as a logical exception. This approach ensures more robust and accurate state tracking, while generalizing naturally to other implicit or conditional events.

\subsection{Correction Mechanisms} 
\label{sec:correction_mechanism}

Logical exceptions detected by the state-transition model represent violations of the assumed game logic and must be corrected to ensure data quality for downstream analyses. To address these cases, \versa employs rule-based handlers that automatically correct the event by either generating missing events or reordering event sequences. Our framework can handle the following problematic cases that exist in raw event stream data:

\begin{figure}[ht]
    \centering
    \begin{subfigure}[t]{0.42\linewidth}
        \centering
        \includegraphics[width=\textwidth, trim={0cm 0cm 1.5cm 3cm},clip]{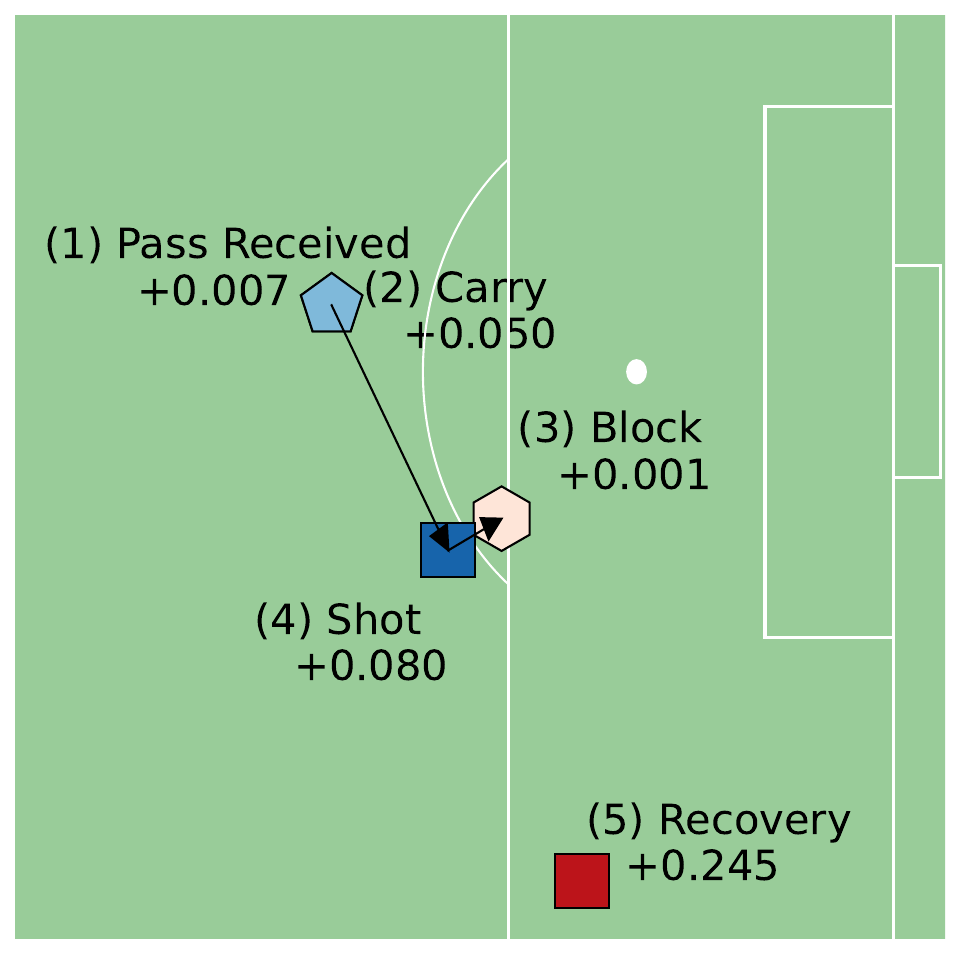}
        \caption{Wrong Sequence: Block $\rightarrow$ Shot}
        \label{fig:wrong_seq}
    \end{subfigure}
    \begin{subfigure}[t]{0.42\linewidth}
        \centering
        \includegraphics[width=\textwidth, trim={0cm 0cm 1.5cm 3cm},clip]{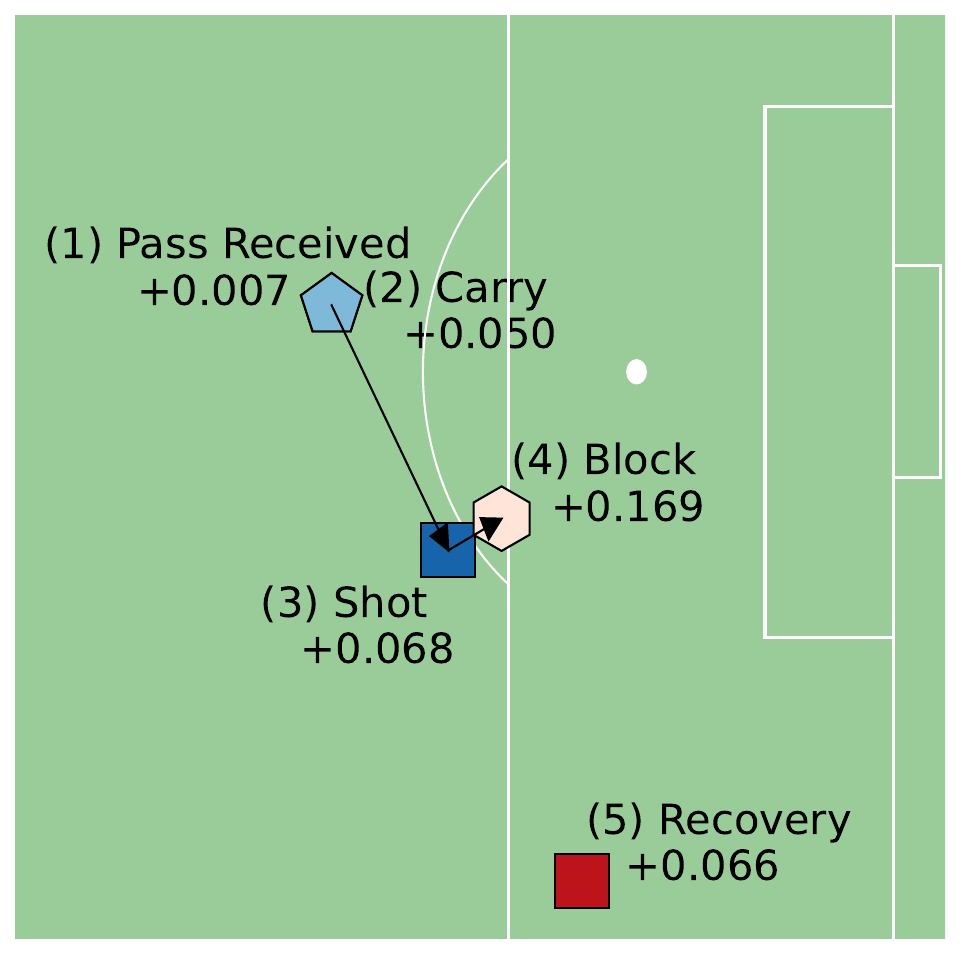}
        \caption{Correct Sequence: Shot $\rightarrow$ Block}
        \label{fig:correct_seq}
    \end{subfigure}
    \caption{This case illustrates an incorrect event order scenario in which a Block event appears before a Shot event}
    \label{fig:case_study_verification}

    \begin{subfigure}[t]{0.42\linewidth}
        \centering
        \includegraphics[width=\textwidth, trim={2.5cm 1cm 0cm 3.5cm},clip]{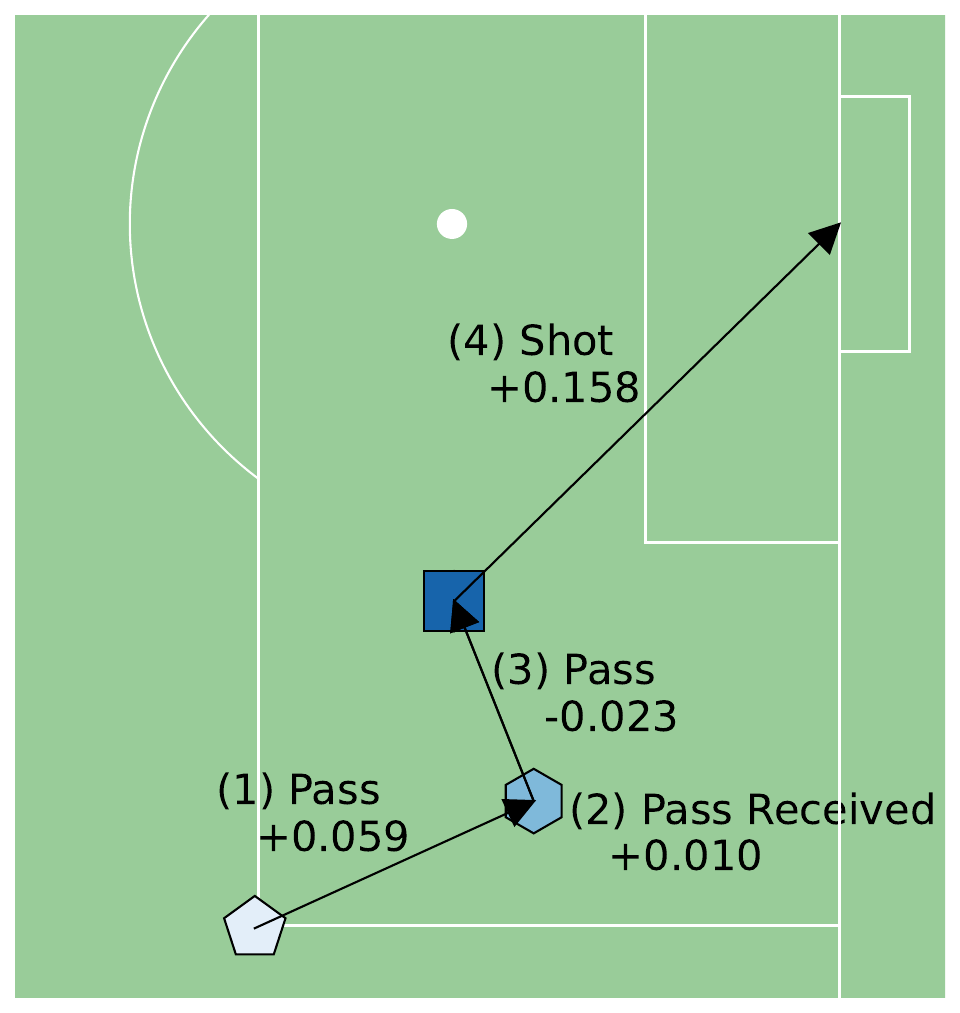}
        \caption{Missing Event: Pass Received}
        \label{fig:missing}
    \end{subfigure}
    \begin{subfigure}[t]{0.42\linewidth}
        \centering
        \includegraphics[width=\textwidth, trim={2.5cm 1cm 0cm 3.5cm},clip]{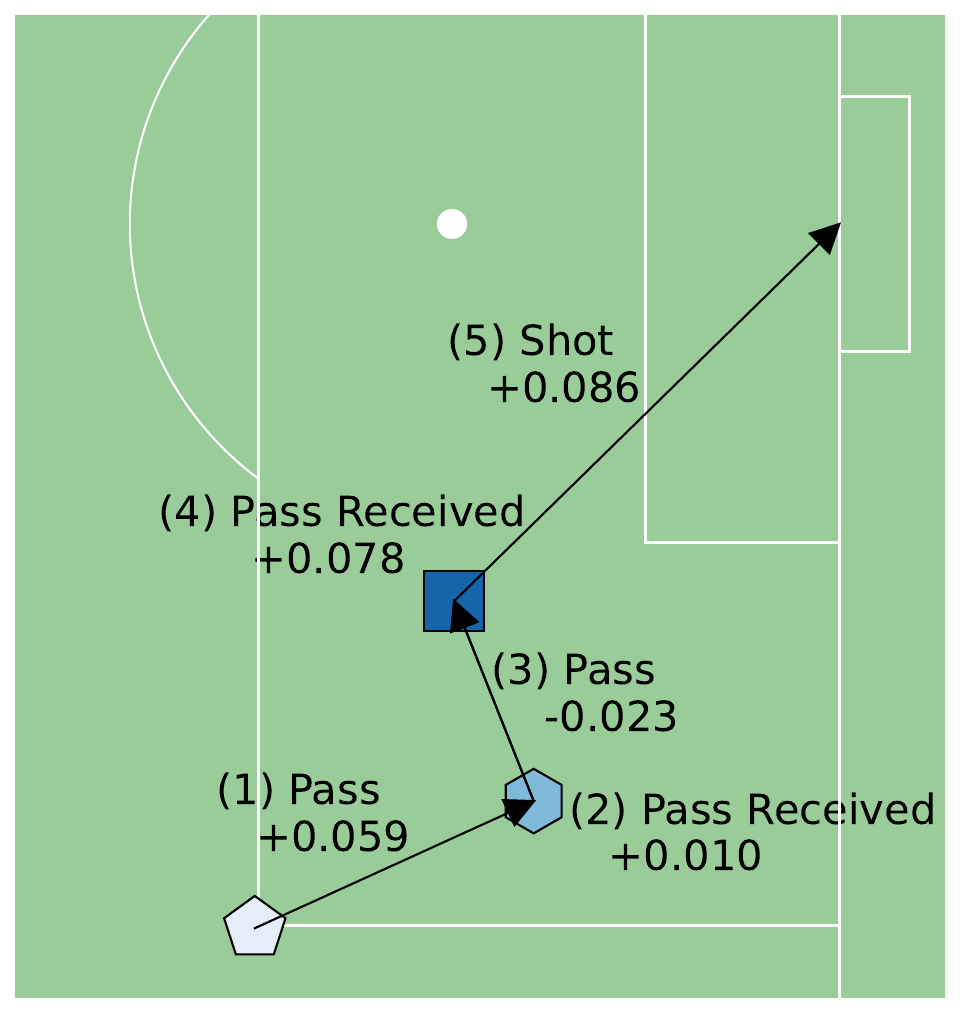}
        \caption{Included Event: Pass Received}
        \label{fig:included}
    \end{subfigure}
        \caption{This case illustrates an incorrect event order scenario in which a shot event without an explicitly recorded pass reception}
        \label{fig:case_study2}
\end{figure}

\begin{enumerate}[leftmargin=*]

\item {\bf Case 1 (Figure~\ref{fig:wrong_seq})} Events appearing in the wrong order (e.g., Block $\to$ Shot): A Block is a defensive action that tries to change the trajectory of the ball, particularly one with a high probability of scoring. 
However, there are some cases where a Block is recorded before a Shot. Since this prevents the Block from being interpreted as a defensive action against the Shot, we cannot properly assess the defensive contributions of the Block events. 
To address the problem, our handler analyzes exceptional cases detected by the verification model and corrects these abnormal event sequences. 
For example, if events are recorded in the order Carry $\to$ Block $\to$ Shot, our model can detect that a Block cannot occur in the `In possession' state and automatically corrects the sequence to ensure the Block occurs after the Shot.

\item {\bf Case 2 (Figure~\ref{fig:missing})} Missing events (e.g., Pass $\to$ Carry or Shot):
If a player performs a Shot action after a successful pass from the same team is played, then there should be a Pass Received event between the Pass and Shot events. However, it is often the case that the receiver attribute is missing, and the reception event is recorded even before the Pass event due to an annotation error.
To address the problem, our model can detect such a situation because the Shot event is allowed to occur only in the `In possession' state instead of the `In transition' state. In this specific example, our handler automatically generates the missing Pass Received event between the two events, ensuring the entire sequence becomes valid.

\end{enumerate}

By integrating transition rules and correction mechanisms, the proposed verification model systematically detects logical exceptions in the event stream and, when necessary, provides a handler that automatically corrects them. Ultimately, this approach not only enhances the reliability of analysis results but also contributes to capturing and evaluating missing events more accurately. For a more detailed description of the algorithm, please refer to Algorithm~\ref{alg:pseudo_code}.

\begin{algorithm}[ht!]
    \caption{State-Transition Model-based Event Data Verification Algorithm}
    \begin{algorithmic}[1]
        \State \textbf{Input:} 
        \Statex \quad - Event dataset $D = \{ e_1, e_2, \dots, e_N \}$ 
        \Statex \quad - All possible states $\mathcal{S} = 
        \{\mathtt{In\ transition, In\ possession}, \dots\}$.
        \Statex \quad - Transition rules $\mathcal{T} = \{ \tau_{(s,a)} \mid (s,a) \in (\mathcal{S} \times A) \}$, where $A$ is the set of all possible actions, and each $\tau_{(s,a)}$ is defined as $(\mathtt{target}, \mathtt{conditions})$.
        \Statex \quad - Handlers $\mathcal{H} = \{ h_{(s, a)} \mid (s, a) \in (\mathcal{S} \times A) \}$.

        \State \textbf{Output:} Verified event dataset $D'$
        
        \State Initialize state machine $\mathcal{M}$ with:
        \Statex \quad - State set $\mathcal{S}$
        \Statex \quad - Initial state: \texttt{Kick-Off}
        \Statex \quad - Transition set $\mathcal{T}$

        \State $D' \gets D$ \Comment{Initialize $D'$ as a copy of $D$}
        
        \For{$i \gets 1$ to $N$}
            \State $s_i \gets \mathcal{M}.state$
            \State $a_i \gets e_i.\text{action}$
            \State $w_i \gets \{ e_{i-5}, \dots, e_{i+5} \}$ 
            
            \State \textbf{attempt} $\mathcal{M}.\text{trigger}(a_i)$
            \If{$\mathcal{M}.\text{state}$ transitions successfully} \Comment{\texttt{target} exists and \texttt{conditions} are met}
                \State $\mathcal{M}.\text{state} \gets \tau_{(s_i, a_i)}.\texttt{target}$
            \Else
                \State $a_i^{\prime}, w_i^{\prime} \gets h_{(s_i, a_i)}(\mathcal{M}.\text{state}, w_i)$ 
                \Comment{Fix or reorder event}
                \State Update $D'$ with $w_i^{\prime}$
                \State $\mathcal{M}.\text{state} \gets \tau_{(s_i, a_i^{\prime})}.\texttt{target}$
            \EndIf
        \EndFor

        \State \textbf{return} $D'$
    \end{algorithmic}
    \label{alg:pseudo_code}
\end{algorithm}

\section{Experiments}

\subsection{Data and Preprocessing}\label{Data and Preprocessing}

We train and evaluate our model using three primary data sources: (1) event stream data provided by \bepro Company, which specializes in sports match video and data analysis, collected from the K League 1 and K League 2 (2021-2025 seasons); (2) publicly available event stream data from \sbomb, collected from the J1 League (2024 season), La Liga (2017/2018 season), and 2018 World Cup; and (3) publicly available event stream data from Wyscout, collected from Premier League, La Liga, Bundesliga, Serie A, Ligue 1 for the 2017/2018 season and 2018 World Cup.

The event stream data contains diverse information related to match events: player and team information involved in the event, the location of the event, action types, and the results of the actions. However, integrating these datasets for analysis poses significant challenges due to differences in the definition of action and result, units of location, and time across providers. To address this, we transform all three data sources into the \versa format, ensuring a unified representation for analysis. 

\subsection{Analysis of Detected Exceptions}
\begin{table}[t!]
    \centering
    \caption{Statistics of datasets and exceptional cases detected by the state transition verification model. `Exception' denotes the ratio of exceptional events to the total, while the `Primary Exception' specifies which exception type appears most frequently.}
    \label{tab:error_ratio}
    \input{tables/Table1}
\end{table}

An analysis of detected exceptions for each data provider using a state transition verification model yielded the results shown in Table~\ref{tab:error_ratio}. 
In this analysis, additional events defined by the state transition model (e.g., Goal Miss, Goal Post, Out, etc.) are not considered exceptions. 
The frequency of detected exceptions varied significantly across providers. \wyscout data exhibited the highest number of exceptions, followed by \sbomb, and then \bepro.

In \wyscout data, the vast majority of exceptions are `Pass Received'. Unlike other providers, \wyscout does not explicitly record a `Pass Received' event following a successful pass. Consequently, our model flags this logical inconsistency as a missing event.
In \bepro and \sbomb data, the vast majority of exceptions are `Carry'. For \bepro, exceptions typically arise when a player performs two consecutive actions with spatial separation exceeding 3 meters. Conversely, \sbomb frequently records `Carry' events for very minor movements (less than 3 meters) or even stationary situations. 
Other observed exceptions showed consistent patterns across providers. For instance, in the \sbomb J1 League, defensive actions were a notable source of exceptions, including `Interception' (4.74\%), `Tackle' (1.04\%), `Recovery' (0.72\%), and `Block' (0.03\%).

To address these diverse anomalies, \versa employs rule-based handlers that automatically correct the detected exceptions. A decline in data reliability could significantly impact the accuracy of analyses. This is particularly concerning for models that heavily rely on the sequential nature of event data, such as \vaep~\cite{Decroos19vaep} and GIM~\cite{Liu20GIM}, as it increases the risk of distorted learning outcomes.

\subsection{Cross-Provider Consistency Evaluation}
\begin{table}[t!]
    \centering
    \caption{Cross-provider consistency across different data representations. The left block reports Normalized Edit Similarity, and the right block reports the Pearson correlation of \vaep values between \sbomb and \wyscout for shared matches (La Liga 2017/2018, 2018 World Cup). Both metrics range from 0 to 1, where higher values indicate greater consistency.}
    \label{tab:consistency}
    \input{tables/Table2}
\end{table}

We evaluate how different data representations affect the consistency between event streams from different providers when describing the same matches. To quantify this, we utilize shared datasets from \sbomb\ and \wyscout: La Liga (2017/2018 season) and the 2018 World Cup.

For each match, we process the shared datasets using four formats: \spadl, \aspadl, \versa, and \simplyversa. Here, \simplyversa is a simplified variant of \versa designed to mitigate inconsistencies caused by provider-specific annotation schemes. It merges overly granular action types and removes certain provider-dependent events; for example, `Pass Corner' and `Shot Corner' are unified into a single `Corner' action. 
We then measure cross-provider consistency using two metrics. First, we compute the Normalized Edit Similarity ($S_{\text{edit}}$) over the sequence of action types within each match half. Note that the Normalized Edit Similarity is defined as:
\[
S_{\text{edit}}(x, y) = 1 - \frac{d_{\text{edit}}(x, y)}{\max(|x|, |y|)},
\]
where $d_\text{edit}$ denotes the edit-distance between two sequences $x$ and $y$. 

Second, we calculate the Pearson correlation of \vaep values, computed based on the sum of \vaep values within each match half, to assess the stability of player evaluation metrics. The results are presented in Table~\ref{tab:consistency}.

The results demonstrate clear differences in cross-provider consistency depending on the data representation used. \spadl and \aspadl show relatively lower similarity, indicating that these formats do not fully ensure logical consistency. In contrast, \versa and \simplyversa achieve higher consistency by enforcing explicit state transitions. This trend is mirrored in the Pearson correlation analysis, where \versa and \simplyversa demonstrate significantly stronger alignment in player valuation metrics compared to the baselines.

\subsection{Transferability to Different Datasets} 

\begin{table}[t!]
    \centering
    \caption{Inference performance of the \vaep model trained via CatBoost using different data formats (\spadl, \aspadl, and \versa) across the \bepro, \sbomb, and \wyscout datasets. The table presents key evaluation metrics: AUROC, Log Loss, Brier Score, and ECE (for scoring (left) and conceding (right) predictions). Best values are highlighted in bold.}
    \label{tab:statsbomb_inference}
    \input{tables/Table3}
\end{table}

We compare the performance of \vaep models trained trained via CatBoost with three different data formats to examine their performance and robustness on heterogeneous data. 
The models were trained and evaluated on a combined dataset incorporating data from all three providers. The results are presented in Table~\ref{tab:statsbomb_inference}.
A clear performance hierarchy emerged: \versa substantially outperforms both \spadl and \aspadl across most evaluated metrics and datasets.

In terms of AUROC, \versa consistently achieves the highest scores across all three providers. This indicates a significantly better ability to distinguish between scoring and conceding outcomes compared to the baseline formats.
This advantage extends to prediction accuracy and calibration. \versa consistently produces the lowest (best) Log Loss and Brier Score, signifying more accurate probabilistic predictions. Furthermore, \versa also recorded the lowest Expected Calibration Error (ECE), indicating that the probabilities it generates are the most reliable and well-calibrated.

Notably, this robust performance demonstrates that \versa's ability to correct errors and enforce logical consistency during data validation is critical. It enables the model to maintain high performance even when trained and evaluated on a complex, heterogeneous dataset amalgamating multiple providers. This suggests that applying the \versa technique to refine and validate datasets can significantly enhance a model’s predictive accuracy, ensuring the robustness of performance evaluation metrics computed using the model across different data sources.

\section{Conclusions}

This study introduces \versa, a state-transition-based verification framework that ensures the logical integrity of soccer event streams by automatically correcting anomalous patterns.
Our experiments demonstrate that \versa significantly enhances cross-provider consistency by resolving logical exceptions. Furthermore, we show that the \vaep model trained on \versa yields substantially more accurate and reliable predictions compared to existing formats.

\section*{AI-Generated Content Acknowledgement}

Generative AI tools were utilized in limited stages of this work. Specifically, ChatGPT was employed to assist in drafting and refining parts of the manuscript text, and GitHub Copilot was used during the early development of code. All outputs generated by these tools were critically reviewed and validated by the authors. No generative AI tools were used for data collection, analysis, or interpretation.

\bibliographystyle{splncs04}
\bibliography{soccer}

\newpage

\end{document}

%% file: packages.tex
\usepackage{kotex}
\usepackage{graphicx} 
\usepackage{multirow}
\usepackage[dvipsnames]{xcolor}
\usepackage{pifont}
\usepackage{tikz}
\usepackage{xspace}
\usepackage{enumitem}
\usepackage[T1]{fontenc}
\usepackage{graphicx}
\usepackage{booktabs}
\usepackage{algorithm}
\usepackage{algpseudocode}
\usepackage[misc]{ifsym}
\usepackage{mwe}

\usepackage{subcaption}
\usepackage{supertabular}
\usepackage{comment}
\usepackage{amssymb}
\usepackage{tabularx} 
\usepackage{xltabular}
\usepackage{siunitx}
\usepackage{orcidlink}  

%% file: macro.tex
\newcommand{\corr}{(\Letter)}

\newcommand{\spadl}{\textsc{Spadl}\xspace}

\newcommand{\aspadl}{\textsc{Atomic-Spadl}\xspace}

\newcommand{\versa}{\textsc{Versa}\xspace}
\newcommand{\simplyversa}{\textsc{Simplified Versa}\xspace}

\newcommand{\vaep}{\textsc{VAEP}\xspace}

\newcommand{\bepro}{\textsc{Bepro}\xspace}
\newcommand{\sbomb}{\textsc{StatsBomb}\xspace}

\newcommand{\wyscout}{\textsc{Wyscout}\xspace}

\newcommand{\bftab}{\fontseries{b}\selectfont}

%% file: tables/Table1.tex
\renewcommand{\arraystretch}{0.7}
\resizebox{1.0\textwidth}{!}{
\setlength{\tabcolsep}{1pt}
\begin{tabular}{lcccccc}
\toprule
{\bf Provider} & {\bf League} & {\bf Season} & {\bf Match} & {\bf Total} & {\bf Exception} & {\bf Primary Exception} \\ \midrule
\multirow{10}{*}{\bepro} 
& & 2021 & 228 & 633,219 & 17.12{\%} & Carry(12.96{\%}) \\ 
& & 2022 & 228 & 643,685 & 15.54{\%} & Carry(12.55{\%}) \\ 
& K League 1 & 2023 & 228 & 660,224 & 16.75{\%} & Carry(12.88{\%}) \\ 
& & 2024 & 228 & 656,527 & 18.81{\%} & Carry(13.94{\%}) \\ 
& & 2025 & 204 & 550,000 & 15.93{\%} & Carry(13.51{\%}) \\ 
& & 2021 & 182 & 465,970 & 12.71{\%} & Carry(12.55{\%}) \\ 
& & 2022 & 220 & 574,089 & 13.77{\%} & Carry(11.76{\%}) \\ 
& K League 2 & 2023 & 234 & 646,500 & 16.08{\%} & Carry(12.11{\%}) \\ 
& & 2024 & 234 & 652,515 & 18.49{\%} & Carry(13.18{\%}) \\ 
& & 2025 & 252 & 682,723 & 19.79{\%} & Carry(13.69{\%}) \\ \midrule
\multirow{3}{*}
{\sbomb}
& La Liga & 2017 & 36 & 110,449 & 20.92{\%} & Carry(14.79{\%}) \\ 
& World Cup & 2018 & 64 & 182,504 & 19.44{\%} & Carry(12.14{\%}) \\ 
& J1 League & 2024 & 380 & 1,018,905 & 22.04{\%} & Carry(14.61{\%}) \\ \midrule
\multirow{6}{*}
{\wyscout}
& Premier League & 2017 & 380 & 965,173 & 38.41{\%} & Pass Received(29.78{\%}) \\ 
& La Liga & 2017 & 380 & 941,788 & 38.07{\%} & Pass Received(29.87{\%}) \\ 
& Bundesliga & 2017 & 306 & 774,850 & 38.15{\%} & Pass Received(29.58{\%}) \\ 
& Serie A & 2017 & 380 & 980,849 & 38.52{\%} & Pass Received(30.52{\%}) \\ 
& Ligue 1 & 2017 & 380 & 946,140 & 38.02{\%} & Pass Received(29.96{\%}) \\ 
& World Cu & 2018 & 64 & 163,520 & 38.99{\%} & Pass Received(31.02{\%}) \\ 
\bottomrule
\end{tabular}
}

%% file: tables/Table2.tex
\renewcommand{\arraystretch}{0.7}
\setlength{\tabcolsep}{1pt}
\begin{tabular}{lcccccc}
    \toprule
    & \multicolumn{2}{c}{\textbf{Normalized Edit Similarity}} &
    \multicolumn{2}{c}{\textbf{Pearson Correlation of VAEP}} \\
    \cmidrule(lr){2-3} 
    \cmidrule(lr){4-5}
    \textbf{Data Format} &
    \textbf{\shortstack[c]{La Liga\\(2017/2018)}} & \textbf{\shortstack[c]{World Cup\\(2018)}} &
    \textbf{\shortstack[c]{La Liga\\(2017/2018)}} & \textbf{\shortstack[c]{World Cup\\(2018)}} \\
    \midrule
    \spadl          & 0.532 & 0.537 & 0.890 & 0.857 \\
    \aspadl         & 0.660 & 0.661 & 0.947 & 0.936 \\
    \versa          & 0.684 & 0.668 & 0.958 & \bftab{0.949} \\
    \simplyversa    & \bftab{0.769} & \bftab{0.757} &
                      \bftab{0.962} & 0.944 \\
    \bottomrule
\end{tabular}


%% file: tables/Table3.tex

\renewcommand{\arraystretch}{0.7}
\resizebox{1.0\textwidth}{!}{
\setlength{\tabcolsep}{1pt}
\begin{tabular}{llcccc}
\toprule
\textbf{Provider} & \textbf{Data Format} & \textbf{AUROC} & \textbf{Log Loss} & \textbf{Brier Score} & \textbf{ECE}\\ \midrule
\multirow{3}{*}
{\bepro} 
& \spadl & 0.7947 / 0.7579 & 0.0537 / 0.0225 & 0.0104 / 0.0037 & 0.0006 / 0.0005 \\ 
& \aspadl & 0.8617 / 0.8446 & 0.0348 / 0.0103 & 0.0068 / 0.0016 & 0.0003 / \bftab{0.0001} \\
& \versa & \bftab{0.8938 / 0.9010} & \bftab{0.0313 / 0.0079} & \bftab{0.0063 / 0.0013}  & \bftab{0.0003} / 0.0003 \\ \midrule
\multirow{3}{*}
{\sbomb} 
& \spadl & 0.8015 / 0.8242 & 0.0470 / 0.0153 & 0.0088 / 0.0024 & 0.0035 / 0.0015 \\ 
& \aspadl & \bftab{0.8921 / 0.8751} & 0.0354 / {\bftab{0.0075}} & 0.0072 / {\bftab{0.0012}} & \bftab{0.0019 / 0.0005} \\
& \versa & 0.8876 / 0.8736 & {\bftab{0.0304}} / 0.0080 & {\bftab{0.0059}} / 0.0013  & 0.0023 / \bftab{0.0005} \\ \midrule
\multirow{3}{*}
{\wyscout} 
& \spadl & 0.7699 / 0.7360 & 0.0662 / 0.0302 & 0.0131 / 0.0051 & 0.0008 / 0.0007 \\ 
& \aspadl & 0.8537 / 0.8484 & 0.0407 / 0.0103 & 0.0080 / 0.0016 & 0.0006 / {\bftab{0.0001}} \\
& \versa & \bftab{0.8789 / 0.8630} &\bftab{0.0352 / 0.0081} & \bftab{0.0070 / 0.0013} & \bftab{0.0005 / 0.0001} \\
\bottomrule
\end{tabular}
}